\documentclass[final,3p,times,twocolumn]{elsarticle}

\usepackage{amssymb}
\usepackage{amsmath}
\usepackage{amsthm}

\usepackage{multirow}
\usepackage{booktabs}
\usepackage[hidelinks]{hyperref}
\usepackage{array}
\usepackage{tabularx}
\usepackage{ragged2e}
\usepackage{threeparttable}
\usepackage[table]{xcolor}
\usepackage{makecell}
\usepackage{adjustbox}

\newcolumntype{L}[1]{>{\RaggedRight\arraybackslash}p{#1}}

\newcolumntype{C}[1]{>{\Centering\arraybackslash}p{#1}}

\newcolumntype{Y}{>{\RaggedRight\arraybackslash}X}

\usepackage{lineno}

\journal{ISPRS Journal of Photogrammetry and Remote Sensing}

\begin{document}

\begin{frontmatter}

\title{SpecTrack: Spectral Prompt Guided Adaptive Experts for Multispectral Object Tracking}

\author[inst1]{Xingyu Tan}
\author[inst1]{Yunrong Qin}
\author[inst1]{Qing Song}
\author[inst1]{Chun Liu}
\author[inst1]{Mengjie Hu\corref{cor1}}

\cortext[cor1]{Corresponding author}

\affiliation[inst1]{organization={Beijing University of Posts and Telecommunications},
            country={China}}


\begin{abstract}
Multispectral image(MSI) and hyperspectral image(HSI) object tracking object tracking exploits recorded band-wise observations to improve target--background discrimination under similar RGB appearance, mixed pixels, illumination variation, occlusion, and clutter. However, existing trackers commonly process all search regions through a fixed capacity spectral--spatial path, ignoring that tracking difficulty varies substantially across frames and target states. Clear regions may require only lightweight local discrimination, whereas ambiguous boundaries and spectrally similar distractors often demand stronger contextual reasoning.
To address this limitation, we propose SpecTrack, a spectral--spatial complexity-aware tracker that formulates MSI tracking as search-region-level adaptive capacity allocation. Its core component, the Spectral Adaptive Mixture-of-Experts (SAMoE) module, provides a capacity-ordered expert pool with progressively increasing latent rank, receptive field, and depth. Expert selection is guided by a Spectral Prompt Router, which fuses semantic context, spatial boundary cues, and a latent channel-variation cue computed after multispectral patch embedding to activate a sparse subset of SAMoE experts for each search region. In parallel, a Shared Global Expert supplies common latent spectral--spatial context to reduce fragmented sparse-routing decisions.  
Experiments on MUST, MSITrack, and HOTC20 demonstrate a favorable accuracy--efficiency trade-off. The accuracy-oriented SpecTrack-L384 achieves state-of-the-art or highly competitive AUCs of 65.2\%, 51.9\%, and 72.6\% on the three benchmarks, while the balanced SpecTrack-B224 reaches 62.4\% AUC at 43.7 FPS on MUST. An additional GOT-10k evaluation indicates RGB-domain architectural generalization, with SpecTrack-L384 achieving 79.3\% AO. The source code will be available at \url{https://github.com/Star-Swift/SpecTrack}.

\end{abstract}




\end{frontmatter}



\section{Introduction}
\label{sec:introduction}

\begin{figure}[t]
    \centering
    \includegraphics[width=1\linewidth]{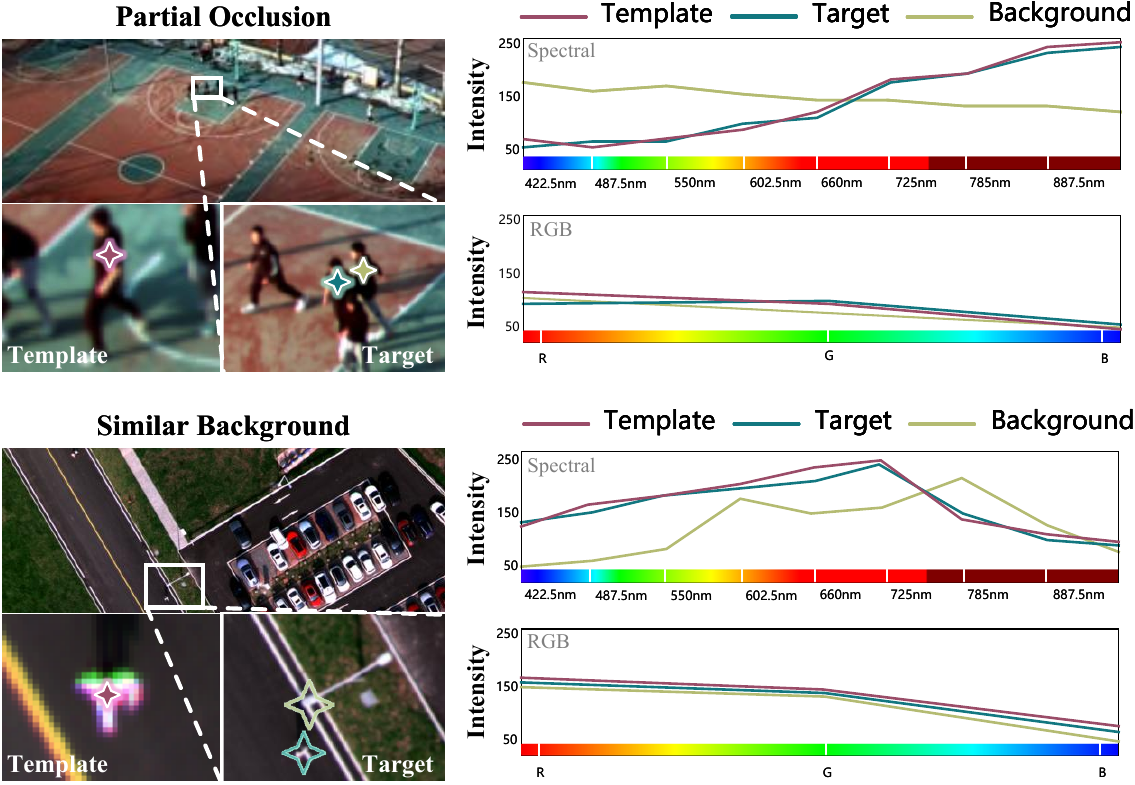}
    \caption{Motivating examples for multispectral tracking. Target and background regions with similar RGB appearance can show separability in recorded band-wise responses. The curves are discrete digital-number or normalized-intensity traces; their physical interpretation depends on sensor response, bandpass, radiometric calibration and acquisition conditions.}
    \label{fig:intro}
\end{figure}

Multispectral and hyperspectral (MSI/HSI) videos provide recorded band-wise observations of a target and its surrounding background. Compared with RGB tracking, these observations offer additional cues for target--background discrimination, especially when the target and distractors share similar RGB appearance but respond differently across the recorded bands. This property is valuable in UAV and remote-sensing tracking scenarios with camouflage, low contrast, illumination variation, occlusion, and cluttered backgrounds. Fig.~\ref{fig:intro} illustrates this motivation with benchmark examples. RGB-similar target and background regions may exhibit different recorded-band responses, providing useful tracking evidence beyond RGB appearance. 

Despite these advantages, MSI/HSI tracking remains challenging because spectral--spatial ambiguity varies substantially across frames and target states. Clear target regions may be localized with lightweight local discrimination, whereas mixed pixels, ambiguous boundaries, partial occlusion, low resolution, illumination variation, and spectrally similar distractors often require stronger contextual reasoning. Prior MSI/HSI tracking studies have improved spectral--spatial representation through band-attention modeling in BAE-Net~\cite{li_bae-net_2020}, ensemble-style feature fusion in SEE-Net~\cite{Li_SEE-Net}, Siamese spectral matching in SiamHYPER~\cite{Liu_SiamHYPER_2022}, dynamic template modeling in SPIRIT~\cite{chen_spirit_2024}, and Transformer-based interaction in UNTrack~\cite{qin_must_2025}. However, these methods usually process different search regions with a similar modeling budget. This fixed-capacity assumption is mismatched with tracking videos, where the difficulty of each search region changes continuously over time.
This observation motivates a different question: \textbf{How can a multispectral tracker use recorded-band-induced target--background separability to allocate appropriate spectral--spatial modeling capacity for search regions with different ambiguity levels?} 

We therefore formulate MSI/HSI tracking as spectral--spatial complexity-aware capacity allocation, where different search regions are assigned different modeling budgets according to their ambiguity. SpecTrack implements this idea with three coupled modules: the \textbf{Spectral Adaptive Mixture-of-Experts (SAMoE)} module provides a capacity-ordered expert pool; the \textbf{Spectral Prompt Router} selects a sparse subset of experts using semantic context, spatial boundary response, and latent channel-variation cues; and the \textbf{Shared Global Expert} supplies common spectral--spatial context to stabilize sparse routing.

Recent MoE-style trackers, including MoETrack~\cite{tang2024revisiting}, SPMTrack~\cite{cai2025spmtrack}, and HotMoE~\cite{sun2025hotmoe}, have shown the potential of expert-based adaptation in visual tracking. In contrast, SpecTrack studies recorded-band-aware capacity allocation for MSI/HSI tracking, where latent channel variation after multispectral patch embedding provides search-region-level routing evidence. We further examine whether this routing behavior is meaningful through mechanism-aligned ablations and diagnostics.

We validate SpecTrack on MUST~\cite{qin_must_2025}, MSITrack~\cite{feng_msitrack_2025} and the 16-band HOTC20 benchmark~\cite{xiong_material_MHT_2020}. GOT-10k~\cite{huang2021GOT10k} is reported as an RGB one-shot architectural generalization test. The main contributions are summarized as follows:
The main contributions are summarized as follows:
\begin{itemize}
    \item We formulate MSI/HSI single object tracking as spectral--spatial complexity-aware capacity allocation, where search regions with different ambiguity levels are assigned different modeling budgets rather than processed by a fixed-capacity path.

    \item We propose the SAMoE module, a capacity-ordered expert pool for MSI/HSI tracking. Its experts differ in latent rank, receptive field, and depth, enabling lightweight modeling for clear regions and stronger reasoning for ambiguous regions.

    \item We design a Spectral Prompt Router and a Shared Global Expert for stable sparse expert routing. The router selects experts from semantic, spatial, and latent channel-variation cues, while the shared expert provides common spectral--spatial context to reduce fragmented local decisions.

    \item We validate SpecTrack on MUST, MSITrack, and HOTC20 with mechanism-aligned ablations, routing diagnostics, and accuracy--efficiency analyses. GOT-10k is additionally used to examine RGB-domain architectural generalization.
\end{itemize}


\section{Related Work}
\label{sec:related_work}

\begin{figure*}[!t]
    \centering
    \begin{minipage}[t]{0.235\textwidth}
        \centering
        \includegraphics[width=\linewidth]{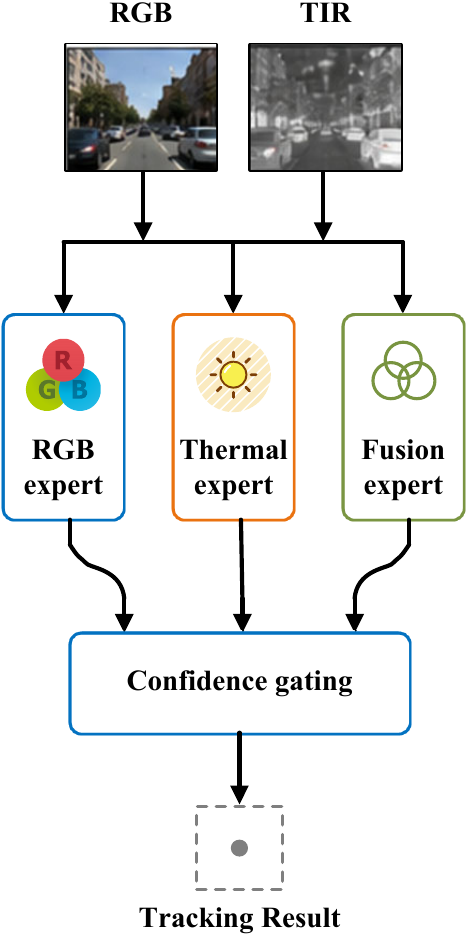}\\[-1mm]
        {\footnotesize (a) MoETrack}
    \end{minipage}\hfill
    \begin{minipage}[t]{0.235\textwidth}
        \centering
        \includegraphics[width=\linewidth]{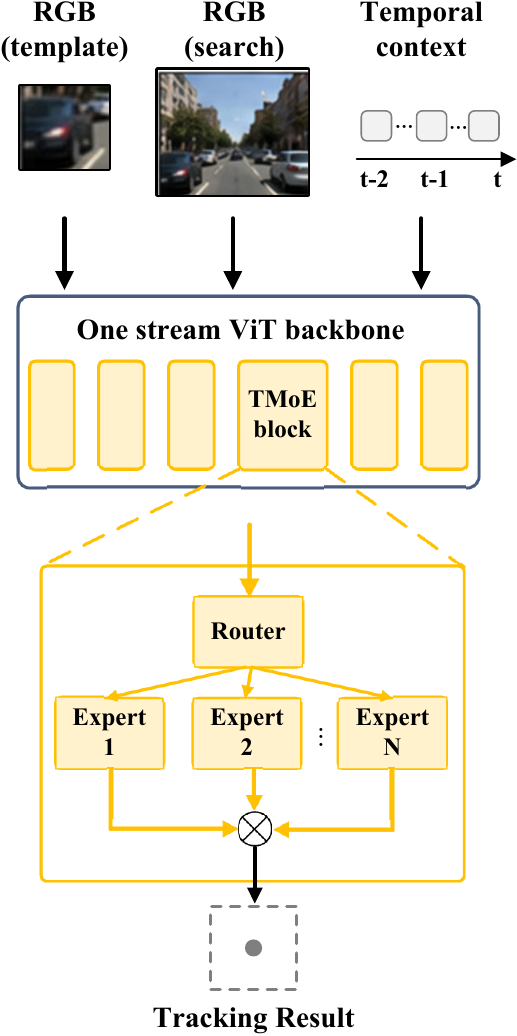}\\[-1mm]
        {\footnotesize (b) SPMTrack}
    \end{minipage}\hfill
    \begin{minipage}[t]{0.235\textwidth}
        \centering
        \includegraphics[width=\linewidth]{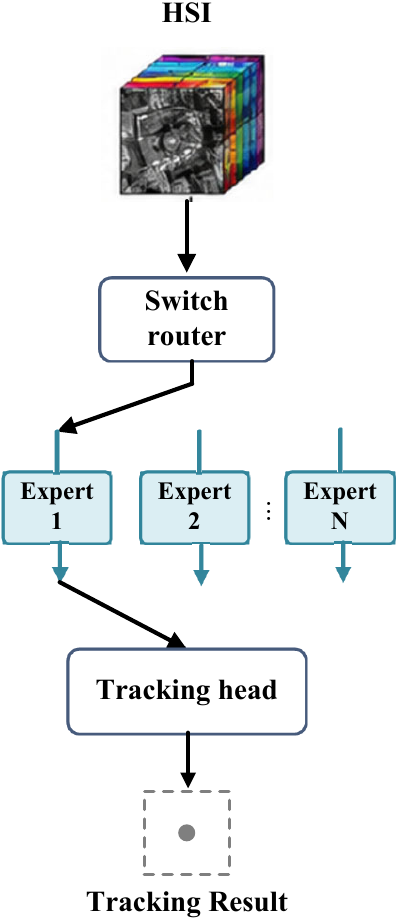}\\[-1mm]
        {\footnotesize (c) HotMoE}
    \end{minipage}\hfill
    \begin{minipage}[t]{0.235\textwidth}
        \centering
        \includegraphics[width=\linewidth]{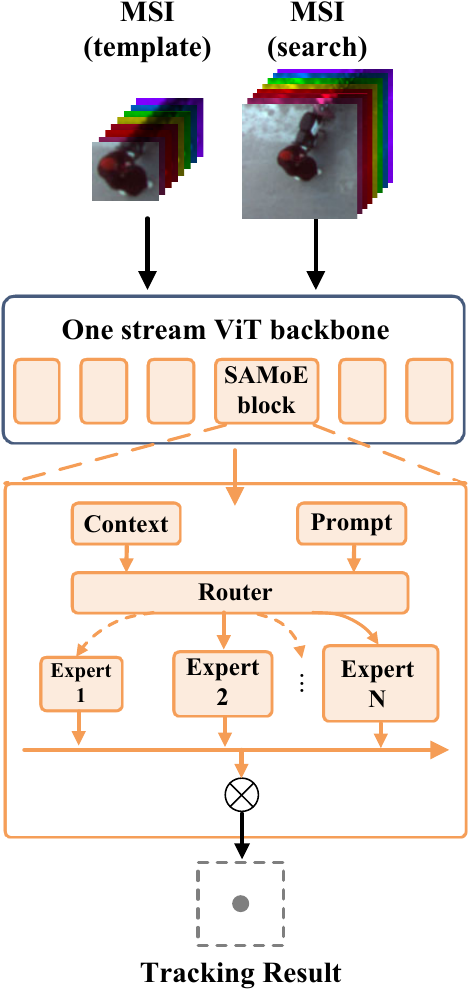}\\[-1mm]
        {\footnotesize (d) SpecTrack}
    \end{minipage}
    \caption{Mechanism level comparison of MoE-style trackers. MoETrack focuses on RGB-T modality validity and confidence-aware fusion; SPMTrack uses tracking-specific MoE blocks for RGB spatio-temporal relation modeling; HotMoE performs sparse HSI expert activation. SpecTrack differs by using recorded-band-induced ambiguity cues to route each MSI/HSI search region to capacity ordered experts inside a shared one-stream backbone, while a Shared Global Expert preserves common latent context.}
    \label{fig:related_positioning}
\end{figure*}

\subsection{Spectral--Spatial Modeling for MSI/HSI Tracking}

MSI/HSI tracking exploits recorded band-wise observations to improve target--background discrimination under low illumination, camouflage, similar color and cluttered background. Early hyperspectral trackers relied on handcrafted spectral descriptors, material unmixing, or correlation-filter formulations~\cite{xiong_material_MHT_2020,Uzkent_2017_CVPR_Workshops}. Deep trackers then introduced band attention, ensemble fusion and cross-band feature interaction, including BAE-Net~\cite{li_bae-net_2020}, SST-Net~\cite{Li_SST-Net_2021}, SEE-Net~\cite{Li_SEE-Net} and CBFF-Net~\cite{gao_cbff_2023}. Recent methods such as SiamHYPER~\cite{Liu_SiamHYPER_2022}, SPIRIT~\cite{chen_spirit_2024}, SENSE~\cite{chen_sense_2024}, MVP-HOT~\cite{zhao_mvphot_2024} and ProFiT~\cite{chen_profit_2025} further show that spectral information should be modeled beyond naive channel expansion.

These studies mainly address how to extract or fuse spectral--spatial representations. SpecTrack studies a complementary problem: how recorded-band-induced evidence can decide the amount of feature-transformation capacity required by each search region. This distinction is important for tracking, because the spectral--spatial ambiguity varies across frames. Clear targets may only require low-cost local modeling, whereas mixed pixels, ambiguous boundaries, occlusion, illumination changes and spectrally similar distractors require stronger contextual modeling. Therefore, SpecTrack treats multispectral information not only as an input feature source, but also as routing evidence for search-region-level capacity allocation.

\subsection{Adaptive Computation and MoE-style Tracking}

MoE and conditional computation have been widely used for adaptive representation learning~\cite{aljundi_expert_2017,riquelme2021scaling,puigcerver_sparse_2024}. Recent tracking methods also introduce MoE-style designs, but they address different adaptation problems. MoETrack~\cite{tang2024revisiting} studies RGB-T tracking under modality validity variation. It uses modality-related experts and confidence-aware fusion to decide when RGB, thermal, or fused predictions should dominate. SPMTrack~\cite{cai2025spmtrack} inserts tracking-specific MoE blocks into an RGB one-stream tracker to improve generic spatio-temporal relation modeling. HotMoE~\cite{sun2025hotmoe} explores sparse expert activation for HSI tracking, where a switch-style router activates suitable HSI expert models to reduce inference cost.

Fig.~\ref{fig:related_positioning} summarizes these mechanism-level differences. The key distinction is the source and role of the routing evidence. MoETrack uses modality validity as the main cue for RGB-T decision fusion. SPMTrack uses visual relation features to improve scalable RGB tracking. HotMoE uses HSI features for sparse expert activation. In contrast, SpecTrack asks whether recorded MSI/HSI evidence inside a shared tracking backbone can estimate the spectral--spatial ambiguity of each search region and then allocate a suitable modeling budget.

This leads to three non-interchangeable design choices. First, SpecTrack uses a capacity-ordered expert ladder rather than homogeneous relation experts or independent modality heads. The experts are organized from low to high capacity through rank, receptive field and depth, so the router can assign lightweight experts to clear regions and stronger experts to ambiguous regions. Second, SpecTrack routes by a Spectral Prompt Router that combines semantic context, spatial boundary response and latent channel-variation cues after multispectral patch embedding. This differs from modality-validity routing in RGB-T tracking and from generic RGB relation routing. Third, SpecTrack introduces a Shared Global Expert to provide common latent context before aggregating sparse local experts, reducing fragmented decisions caused by local sparse routing.

A direct transplant of MoETrack, SPMTrack, or HotMoE is not fully equivalent because these methods are designed for different inputs, expert outputs, backbones and training protocols. To avoid conflating the proposed idea with implementation differences, we complement published-method comparisons with same-backbone mechanism controls in Sec.~\ref{sec:ablation}. These controls approximate the core alternatives under the same SpecTrack-B224 backbone, input resolution, training split and evaluation protocol: homogeneous MoE-style experts test whether multiple experts alone are sufficient; semantic-only routing tests generic visual routing; capacity experts without prompt routing test whether capacity diversity alone is enough; and local experts without the Shared Global Expert test whether sparse HSI-style expert activation can replace shared context modulation.

\subsection{Remote-Sensing Interpretation Scope}
Physical interpretation in remote sensing depends on sensor metadata, bandpass, radiometric calibration, illumination and acquisition geometry. Tracking benchmarks often provide digital numbers or normalized intensities. SpecTrack therefore uses qualitative spectral curves as recorded-band separability illustrations and uses latent channel variation as a routing descriptor after patch embedding. When calibrated reflectance and sensor response functions are available, the same routing idea can be extended toward sensor-aware material cues.

\begin{figure*}[!t]
    \centering
    \includegraphics[width=1\linewidth]{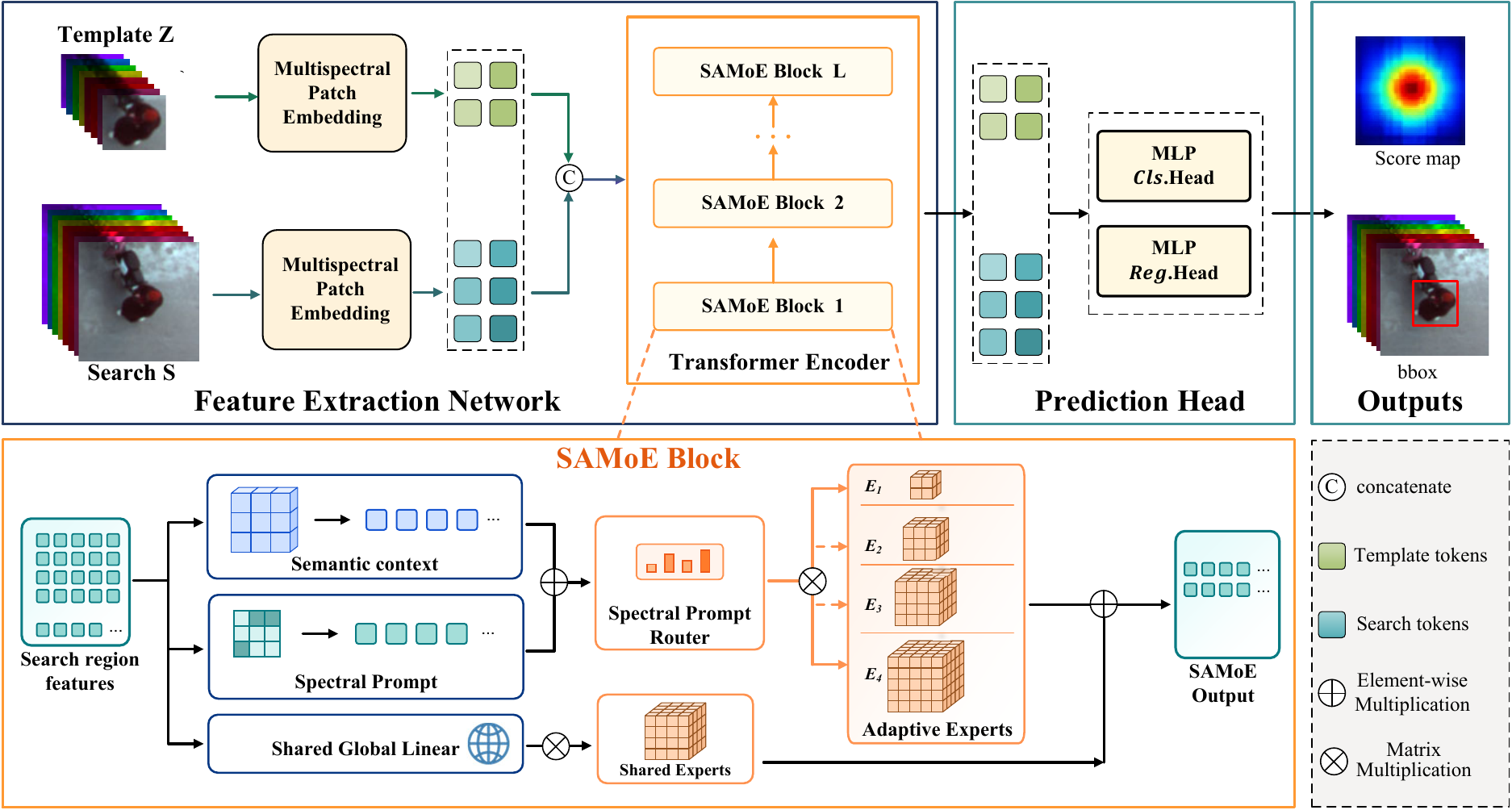}
    \caption{Overall framework of SpecTrack. Template and search MSI inputs are patch embedded and interact in a one-stream Transformer encoder. Standard blocks process the early encoder stages, while later blocks use SAMoE, which contains an adaptive expert set, a Spectral Prompt Router and a Shared Global Expert. The center head consumes the final search tokens and predicts the score map, target size, offset and bounding box.}
    \label{fig:framework}
\end{figure*}

\section{Method}
\label{sec:method}

\subsection{Framework Overview}
\label{sec:framework}
As illustrated in Fig.~\ref{fig:framework}, SpecTrack follows a one-stream tracking paradigm and reorganizes the later encoder blocks around the spectral--spatial difficulty formulation. Template and search tokens interact in the shared encoder, while the prediction head consumes the final search-region tokens for localization. The core idea is to allocate expert capacity at the search-region feature level for heterogeneous search conditions.

The architecture has three roles. First, the capacity-ordered expert set provides a low-to-high representation ladder for search regions with different ambiguity levels. Second, the Spectral Prompt Router converts semantic, spatial and latent channel-variation cues into a sample-level Top-$k$ expert selection. Third, the Shared Global Expert supplies common latent context before the sparse expert update is inserted back into the search tokens. These operations are performed after patch embedding in latent feature space; SAMoE is an adaptive feature-transformation block guided by recorded-band-induced cues.

For a SAMoE block at layer $l$, let $\hat{\mathbf{Z}}^l_s\in\mathbb{R}^{B\times L_s\times d}$ denote the search tokens after the standard self-attention sublayer. We reshape them into a latent search feature map $x=\mathcal{U}(\hat{\mathbf{Z}}^l_s)\in\mathbb{R}^{B\times d\times H\times W}$, where $L_s=HW$ and $\mathcal{U}(\cdot)$ denotes token-to-map unflattening. SAMoE computes one residual update for the whole search sample and then flattens it back to the search token positions. The router is therefore sample conditioned at the search-region level.

\paragraph{Feature extraction network}
The network takes a template image $Z \in \mathbb{R}^{H_z \times W_z \times C_b}$ from the first frame and a search region $S \in \mathbb{R}^{H_s \times W_s \times C_b}$ from the current frame, where $C_b$ denotes the number of recorded spectral bands. The multispectral input is partitioned into patches and projected into a shared latent space:
\begin{equation}
    e_i = \mathrm{PE}(p_i), \qquad p_i \in \mathbb{R}^{P \times P \times C_b},
\end{equation}
where $\mathrm{PE}(\cdot)$ denotes the patch projection layer. Template and search embeddings are concatenated, supplemented with positional encodings and processed by the one-stream Transformer encoder.

\paragraph{SAMoE block}
In later encoder blocks, SpecTrack inserts SAMoE after self-attention and before the feed-forward network. The block is written as:
\begin{equation}
    \hat{\mathbf{Z}}^l = \mathbf{Z}^{l-1} + \mathrm{MSA}(\mathrm{LN}(\mathbf{Z}^{l-1})),
\end{equation}
\begin{equation}
    \Delta x^l_s = \mathrm{SAMoE}(\mathcal{U}(\hat{\mathbf{Z}}^l_s)),
\end{equation}
\begin{equation}
    \tilde{\mathbf{Z}}^l = \hat{\mathbf{Z}}^l + \alpha_l\,\mathcal{R}_s(\mathcal{F}(\Delta x^l_s)),
\end{equation}
\begin{equation}
    \mathbf{Z}^l = \tilde{\mathbf{Z}}^l + \mathrm{FFN}(\mathrm{LN}(\tilde{\mathbf{Z}}^l)),
\end{equation}
where $\mathcal{F}(\cdot)$ flattens a feature map back to tokens, $\mathcal{R}_s(\cdot)$ inserts the update only into the search token positions and $\alpha_l$ is a learnable residual scale. SAMoE updates the search tokens in this branch, while template tokens influence the search representation through the preceding one-stream attention.

\paragraph{Prediction head}
The prediction head receives the final search region tokens and reshapes them to recover the two-dimensional feature map. Following OSTrack~\cite{Ye2022OSTrack}, we use a center point prediction head in which classification and regression branches estimate the score map, target size, offset and final bounding box.

\subsection{Spectral Prompt Router}
\label{sec:router_detail}
Effective expert assignment is critical for adaptive computation. Spatial semantics alone may be insufficient for camouflaged targets, mixed pixels, ambiguous boundaries and spectrally similar distractors. The Spectral Prompt Router therefore combines semantic context with a fixed spatial high-pass prompt and a latent channel-variation prompt.

Given $x\in\mathbb{R}^{B\times d\times H\times W}$, the router first normalizes the feature and obtains a semantic descriptor:
\begin{equation}
    \bar{x}=\mathrm{LN}(x), \qquad z=\mathrm{GAP}(\bar{x})\in\mathbb{R}^{B\times d}.
    \label{eq:router_semantic}
\end{equation}
The spatial prompt is computed by a fixed depth-wise Laplacian filter:
\begin{equation}
    K_{lap}=\begin{bmatrix}0&-1&0\\-1&4&-1\\0&-1&0\end{bmatrix}, \qquad h_{spa}=\mathrm{DWConv}_{K_{lap}}^{rep}(\bar{x}),
    \label{eq:spatial_laplacian}
\end{equation}
where $rep$ denotes replicate padding at image boundaries. We pool the absolute high-pass response and normalize it by the average feature magnitude:
\begin{equation}
    e_{spa}=\frac{\mathrm{GAP}(|h_{spa}|)}{\mathrm{GAP}(|\bar{x}|)+\tau}\in\mathbb{R}^{B\times d},
    \label{eq:spatial_prompt}
\end{equation}
where $\tau$ is a small numerical constant.

The spectral prompt summarizes latent channel variation after multispectral patch embedding. The patch projection is initialized by spectral-aware channel expansion and optimized on MSI/HSI tracking data, so the resulting latent channels carry recorded-band-induced feature contrast. A first-order finite difference compactly describes this contrast for expert routing:
\begin{equation}
    (D_c\bar{x})_{b,c,h,w}=\begin{cases}
    0, & c=1,\\
    \bar{x}_{b,c,h,w}-\bar{x}_{b,c-1,h,w}, & c>1,
    \end{cases}
    \label{eq:channel_diff}
\end{equation}
followed by the same magnitude normalization:
\begin{equation}
    e_{spe}=\frac{\mathrm{GAP}(|D_c\bar{x}|)}{\mathrm{GAP}(|\bar{x}|)+\tau}\in\mathbb{R}^{B\times d}.
    \label{eq:spectral_prompt}
\end{equation}
This operation provides a compact recorded-band-induced latent contrast cue for routing capacity in the learned feature space.

The two prompts are concatenated and fused by a lightweight two-layer projection:
\begin{equation}
    e_{prompt}=\phi_f([e_{spa},e_{spe}])=W_2^f\,\delta\left(W_1^f\,\mathrm{LN}([e_{spa},e_{spe}])\right),
    \label{eq:prompt_fusion}
\end{equation}
where $[\cdot,\cdot]$ denotes channel concatenation and $\delta$ is GELU. The routing logits are:
\begin{equation}
    r=W_s z+W_f e_{prompt}+\epsilon, \qquad r\in\mathbb{R}^{B\times N}.
    \label{eq:routing_logit_detail}
\end{equation}
During training, $\epsilon_{b,i}\sim\mathcal{N}(0,\sigma_{\epsilon}^{2})$ is a small Gaussian routing jitter with a fixed code-default standard deviation. It stabilizes early routing and uses the same code default across benchmarks. During validation and inference, $\epsilon$ is set to zero.

The final sparse gate is computed from the softmax probabilities $p=\mathrm{Softmax}(r)$. Let $\mathcal{I}_b=\mathrm{TopK}(p_b,k)$ be the selected expert indices of sample $b$. The gate is renormalized over the selected experts:
\begin{equation}
    g_{b,i}=\begin{cases}
    \dfrac{p_{b,i}}{\sum_{j\in\mathcal{I}_b}p_{b,j}}, & i\in\mathcal{I}_b,\\[6pt]
    0, & i\notin\mathcal{I}_b.
    \end{cases}
    \label{eq:topk_gate_detail}
\end{equation}
Thus, unselected experts have zero gate value and are skipped during inference. The gate is sample conditioned at the search-region level and shared by all tokens in the current search region. This definition also explains the routing diagnostics in the experiments: usage counts selected Top-$k$ slots, whereas gate weight records the normalized selected $g_{b,i}$ values.

\subsection{Adaptive Expert Set}
\label{sec:samoe_detail}
Standard MoE architectures usually use homogeneous experts with identical structures. SAMoE uses a capacity-ordered expert set whose experts are organized by computational and representational capacity.

\paragraph{Capacity schedule}
For an input feature $x\in\mathbb{R}^{B\times d\times H\times W}$, SAMoE contains $N$ experts $\mathcal{E}=\{E_i\}_{i=1}^{N}$ from lightweight to strong. Expert $E_i$ first projects $d$ channels to a rank $r_i$, applies $t_i$ repeated latent spatial-frequency mixer units and projects the feature back to $d$ channels:
\begin{equation}
    d \xrightarrow{W_i^{in}} r_i \xrightarrow{\text{$t_i$ mixer units}} r_i \xrightarrow{W_i^{out}} d .
    \label{eq:expert_bottleneck}
\end{equation}
The nested rank schedule used in the main experiments is:
\begin{equation}
    r_i = \max \left(r_{\min},\left\lfloor \frac{i}{N}r_{\max}\right\rfloor\right), \qquad i=1,\ldots,N,
    \label{eq:nested_rank}
\end{equation}
where $r_{\min}$ and $r_{\max}$ are fixed by the backbone width and are kept unchanged across datasets. The expert depth is scaled linearly:
\begin{equation}
    t_i = \max\left(1,\left\lceil \frac{i}{N}t_{\max}\right\rceil\right).
    \label{eq:depth_scale}
\end{equation}
For the default four-expert setting, the resulting expert schedule is summarized in Tab.~\ref{tab:expert_schedule}. This table gives the computation order required to reproduce the block at the architectural level; low-level constants such as initialization seeds, normalization epsilons and PyTorch layer defaults follow the released implementation.

\begin{table*}[!t]
    \centering
    \caption{Default capacity schedule of SAMoE when $N=4$. The rank values are produced by Eq.~\ref{eq:nested_rank}; $w_i$ is the non-overlapping local window size used by the FFT mixer, $\kappa_i$ is the depth-wise convolution kernel size and $t_i$ is the number of repeated rank-space mixer units.}
    \label{tab:expert_schedule}
    \fontsize{8pt}{9pt}\selectfont
    \begin{tabularx}{\linewidth}{@{}C{0.10\linewidth}C{0.10\linewidth}C{0.10\linewidth}C{0.10\linewidth}C{0.10\linewidth}Y@{}}
        \toprule
        Expert & Rank & Window $w_i$ & Kernel $\kappa_i$ & Depth $t_i$ & Intended role \\
        \midrule
        $E_1$ & $r_1$ & $4\times4$ & $3\times3$ & 1 & Low-cost local response for clear targets \\
        $E_2$ & $r_2$ & $8\times8$ & $5\times5$ & 2 & Moderate local context and boundary modeling \\
        $E_3$ & $r_3$ & $16\times16$ & $7\times7$ & 3 & Larger spectral-spatial ambiguity modeling \\
        $E_4$ & $r_4$ & $32\times32$ & $9\times9$ & 4 & Highest capacity for difficult search regions \\
        \bottomrule
    \end{tabularx}
\end{table*}

\paragraph{Expert computation path}
Each local expert has its own parameters; local projection, convolution, FFT mixing, MLP and output projection weights are independent across experts. Given $x$, expert $E_i$ first applies normalization, a $1\times1$ rank projection and a depth-wise convolution:
\begin{equation}
    u_i^{(0)} = \delta\left(\mathrm{DWConv}_{\kappa_i}\left(W_i^{in}\,\mathrm{LN}(x)\right)\right),
    \label{eq:expert_preproj}
\end{equation}
where $\delta(\cdot)$ is GELU. The convolution is applied in the spatial domain and injects local context before the FFT mixer.

For the $m$-th mixer unit, $u_i^{(m-1)}$ is split into non-overlapping spatial windows of size $w_i\times w_i$:
\begin{equation}
    \{u_{i,j}^{(m-1)}\}_{j=1}^{J_i}=\mathcal{P}_{w_i}\left(u_i^{(m-1)}\right).
    \label{eq:window_partition}
\end{equation}
If $H$ or $W$ is not divisible by $w_i$, replicate padding is applied before partitioning and the padded area is cropped after window reversal. Within each window, $Q$, $K$ and $V$ are produced by a linear projection in the rank space:
\begin{equation}
    (q_{i,j},k_{i,j},v_{i,j})=\mathrm{Split}\left(W_i^{qkv}u_{i,j}^{(m-1)}\right).
    \label{eq:qkv_projection}
\end{equation}
The frequency interaction is computed over the two spatial dimensions of each window:
\begin{equation}
    a_{i,j}=\mathrm{Re}\left(\mathcal{F}^{-1}_{2D}\left(\mathcal{F}_{2D}(q_{i,j})\odot\mathcal{F}_{2D}(k_{i,j})\right)\right),
    \label{eq:fft_interaction}
\end{equation}
\begin{equation}
    o_{i,j}=a_{i,j}\odot v_{i,j}.
    \label{eq:fft_value_modulation}
\end{equation}
The window outputs are reversed to the full feature map and passed through a rank-space MLP with a residual connection:
\begin{equation}
    \bar{u}_i^{(m)}=\mathcal{P}^{-1}_{w_i}\left(\{o_{i,j}\}_{j=1}^{J_i}\right),
\end{equation}
\begin{equation}
    u_i^{(m)}=u_i^{(m-1)}+\bar{u}_i^{(m)}+\mathrm{MLP}_i\left(\mathrm{LN}\left(u_i^{(m-1)}+\bar{u}_i^{(m)}\right)\right).
    \label{eq:expert_inner_residual}
\end{equation}
The local residual update produced by expert $E_i$ is:
\begin{equation}
    \Delta_i(x)=W_i^{out}u_i^{(t_i)} \in \mathbb{R}^{B\times d\times H\times W}.
    \label{eq:expert_output}
\end{equation}
Equations~\ref{eq:expert_preproj}--\ref{eq:expert_output} clarify the operation order: normalization, rank projection, depth-wise convolution, local window partition, $QKV$ projection, two-dimensional spatial FFT interaction on latent features, inverse FFT, value modulation, window reversal, rank-space MLP and output projection.

\subsection{Shared Experts}
\label{sec:shared_detail}
Sparse expert selection reduces redundant computation, while selected experts still require common context to avoid fragmented local updates. The Shared Global Expert provides a common modulation signal for all selected local experts. It uses global pooling followed by a small bottleneck projection:
\begin{equation}
    s_g=\mathrm{GAP}(\mathrm{LN}(x)), \qquad m_g=\sigma\left(W_2^g\,\delta(W_1^g s_g)\right)\in\mathbb{R}^{B\times d},
    \label{eq:shared_global_gate}
\end{equation}
where $\sigma(\cdot)$ is the sigmoid function. The modulation vector is broadcast to the spatial dimensions and applied to each selected expert update:
\begin{equation}
    \tilde{\Delta}_i(x)=\Delta_i(x)\odot \mathrm{Broadcast}(m_g).
    \label{eq:shared_modulation}
\end{equation}
The SAMoE output is then:
\begin{equation}
    \mathrm{SAMoE}(x)=W_o\left(\sum_{i=1}^{N}g_i(x)\tilde{\Delta}_i(x)\right).
    \label{eq:samoe_output_detail}
\end{equation}
The shared branch provides common latent context before the sparse expert outputs are inserted into the backbone residual path.

\subsection{Loss Function}
\label{sec:training_detail}
The tracking loss follows the OSTrack~\cite{Ye2022OSTrack} center point prediction strategy. We use weighted focal loss~\cite{law2018cornernet} for classification and a combination of $\ell_1$ loss and generalized IoU loss~\cite{rezatofighi2019generalized} for regression:
\begin{equation}
    \mathcal{L}_{track}=\mathcal{L}_{cls}+\lambda_G\mathcal{L}_{GIoU}+\lambda_{L1}\mathcal{L}_{L1}.
    \label{eq:tracking_loss_detail}
\end{equation}

The auxiliary routing loss is computed at the sample level. Let $p_{b,i}$ be the pre-TopK probability from the router and let $\mathbb{1}(i\in\mathcal{I}_b)$ indicate whether expert $i$ is selected for sample $b$. The importance and load statistics are:
\begin{equation}
    \mathrm{Imp}_i=\sum_{b=1}^{B}p_{b,i}, \qquad \mathrm{Load}_i=\sum_{b=1}^{B}\mathbb{1}(i\in\mathcal{I}_b).
    \label{eq:imp_load_detail}
\end{equation}
We use the squared coefficient of variation:
\begin{equation}
    CV(v)^2=\frac{\mathrm{Var}(v)}{\mathrm{Mean}(v)^2+\tau}.
    \label{eq:cv_detail}
\end{equation}
Because SAMoE experts have different costs, the default loss also includes a simple complexity bias. The relative complexity proxy of expert $E_i$ is:
\begin{equation}
    \chi_i=r_i\,t_i\,(\kappa_i^2+w_i^2), \qquad c_i=\frac{\chi_i}{\max_j \chi_j+\tau}.
    \label{eq:complexity_proxy}
\end{equation}
The complexity term penalizes excessive reliance on the strongest experts while still allowing them to be selected when the router assigns high probability:
\begin{equation}
    \mathcal{L}_{comp}=\sum_{i=1}^{N}c_i\frac{\mathrm{Imp}_i}{\sum_j\mathrm{Imp}_j+\tau}.
    \label{eq:complexity_loss}
\end{equation}
The final MoE regularizer is:
\begin{equation}
    \mathcal{L}_{moe}=\frac{1}{2}CV(\mathrm{Imp})^2+\frac{1}{2}CV(\mathrm{Load})^2+\lambda_c\mathcal{L}_{comp}.
    \label{eq:moe_loss_detail}
\end{equation}
The total training objective is:
\begin{equation}
    \mathcal{L}=\mathcal{L}_{track}+\lambda_{moe}\mathcal{L}_{moe}.
    \label{eq:final_loss_detail}
\end{equation}
We set $\lambda_G=2$, $\lambda_{L1}=5$ and $\lambda_{moe}=0.01$ unless otherwise specified; the small coefficient $\lambda_c$ follows the same code default in all datasets.

The terminology used in the scaling ablation is defined by the equations above. \textit{Nested rank + max scaling} means that ranks follow Eq.~\ref{eq:nested_rank} and the complexity weight is normalized by the maximum proxy in Eq.~\ref{eq:complexity_proxy}. \textit{Min complexity scaling} uses the same proxy but normalizes by $\min_j\chi_j$, which over-amplifies the cost difference and is used only as an ablation. \textit{Exponential rank scaling} replaces Eq.~\ref{eq:nested_rank} with an exponentially increasing rank sequence while keeping the same window, kernel and depth order. \textit{Without complexity bias} sets $\lambda_c=0$ and \textit{without auxiliary loss} sets $\lambda_{moe}=0$.

During inference, the auxiliary loss and routing noise are disabled. The tracker uses one static template and one dynamically updated template~\cite{yan2021learning}; the online template is updated only when both the fixed interval and confidence threshold criteria are satisfied.

\section{Experiments}
\label{sec:experiments}

\subsection{Implementation Details}

SpecTrack is implemented in Python 3.13 with PyTorch 2.9. Training uses eight NVIDIA RTX 3090 GPUs and speed is measured on one RTX 3090. The default setting uses a Fast-iTPN/OSTrack-style one-stream encoder, $N=4$ capacity-ordered experts and Top $k=2$ routing.

\paragraph{Model}
We develop five variants to examine accuracy--speed trade-offs (Tab.~\ref{tab:spectrack_models}). L variants use HiViT-L~\cite{Zhang2023HiViTAS}, B variants use HiViT-B and T224 uses HiViT-T. The encoders are initialized with FastiTPN~\cite{Tian2024FastiTPNIP} and adapted to MSI/HSI inputs through spectral-aware channel expansion.

\paragraph{Spectral aware channel expansion}
The original patch projection of an RGB pretrained model has weights $W^{rgb}\in\mathbb{R}^{d\times3\times P\times P}$. For an MSI/HSI input with $C$ bands, we construct $W^{C}\in\mathbb{R}^{d\times C\times P\times P}$ from the RGB pretrained projection to preserve a stable initialization. In the eight band MSI setting with band centers $\{422.5,487.5,550.0,602.5,660.0,725.0,785.0,887.5\}$\,nm, we use a spectral aware mapping based on wavelength proximity: bands with $\lambda\le500.0$\,nm are initialized from the blue RGB channel, bands with $500.0<\lambda<620.0$\,nm from the green channel and bands with $\lambda\ge620.0$\,nm from the red channel. Let $m(c)\in\{B,G,R\}$ denote the mapped RGB channel for band $c$ and $\eta_c$ denote an optional scale factor. The expanded kernel is:
\begin{equation}
    W^{C}_{:,c,:,:}=\eta_c W^{rgb}_{:,m(c),:,:}.
    \label{eq:channel_expansion_insert}
\end{equation}
We set $\eta_c=1$ for \textit{insert} and $\eta_c=1/2$ for \textit{insert halfcopy}. For other band layouts, we use interpolation-based initialization:
\begin{equation}
    W^{C}_{:,c,:,:}=\frac{3}{C}\sum_{r=1}^{3}\alpha_{c,r}W^{rgb}_{:,r,:,:},\quad \sum_{r=1}^{3}\alpha_{c,r}=1,
    \label{eq:channel_expansion_interp}
\end{equation}
where $\alpha_{c,r}$ interpolates the band center to RGB reference centers when metadata are available and uses $\alpha_{c,r}=1/3$ otherwise. The factor $3/C$ keeps the activation scale close to the RGB projection. HOTC20 uses this fallback because its 470--620\,nm range is concentrated in the visible spectrum.

\begin{table*}[t]
    \centering
    \caption{Configuration of SpecTrack model variants.}
    \label{tab:spectrack_models}
    \fontsize{8pt}{9pt}\selectfont
    \begin{tabular*}{\textwidth}{@{\extracolsep{\fill}}lcccccc@{}}
        \toprule
        \multirow{2}{*}{Model} & Transformer & Search & Template & Params & FLOPs & Speed \\
        & Encoder & Resolution & Resolution & (M) & (G) & (fps) \\
        \midrule
        SpecTrack-L384 & HiViT-L & 384$\times$384 & 192$\times$192 & 434 & 289 & 5.9 \\
        SpecTrack-L224 & HiViT-L & 224$\times$224 & 112$\times$112 & 434 & 109 & 19.4 \\
        SpecTrack-B384 & HiViT-B & 384$\times$384 & 192$\times$192 & 117 & 81 & 17.2 \\
        SpecTrack-B224 & HiViT-B & 224$\times$224 & 112$\times$112 & 117 & 31 & 43.7 \\
        SpecTrack-T224 & HiViT-T & 224$\times$224 & 112$\times$112 & 30 & 9 & 80.1 \\
        \bottomrule
    \end{tabular*}
\end{table*}

\paragraph{Training protocol}
The main tables use dataset-specific training. MUST, MSITrack and HOTC20 models are trained on their official training splits; GOT-10k follows its one-shot RGB protocol. Template and search images expand target boxes by factors of 2 and 4. We use AdamW, batch size 32, learning rate $10^{-4}$, 25 epochs and step decay at epoch 15.

\paragraph{Inference}
The online template is updated every 25 frames when the confidence score exceeds 0.7. Following standard practice~\cite{Ye2022OSTrack}, we apply a Hann window penalty to incorporate positional priors.

\subsection{Dataset Versions, Protocols and Metrics}

We evaluate SpecTrack on four benchmarks (Tab.~\ref{tab:protocol}): MUST~\cite{qin_must_2025}, MSITrack~\cite{feng_msitrack_2025}, HOTC20~\cite{xiong_material_MHT_2020} and GOT-10k~\cite{huang2021GOT10k}. The first three support MSI/HSI evaluation; GOT-10k is the RGB one-shot generalization setting.

\begin{table}[h]
    \centering
    \caption{Dataset versions and protocols used in the main comparisons. External data and mixed training results should be reported separately.}
    \label{tab:protocol}
    \fontsize{8pt}{9pt}\selectfont
    \begin{tabular*}{\columnwidth}{@{\extracolsep{\fill}}l c c l@{}}
        \toprule
        Dataset & Bands & Split  & Protocol note \\
        \midrule
        MUSTT~\cite{qin_must_2025} & 8 & 160/90 train/test & UAV MSI \\
        MSITrack~\cite{feng_msitrack_2025} & 8 & 220/80 train/test & 395--950\,nm MSI \\
        HOTC20~\cite{xiong_material_MHT_2020} & 16 & 40/35 train/test & 470--620\,nm HSI \\
        GOT-10k~\cite{huang2021GOT10k} & 3 & official train/test & RGB protocol \\
        \bottomrule
    \end{tabular*}
\end{table}

We report standard metrics: Area Under the Curve (AUC), Success Rates at 0.5/0.75 ($SR_{0.5/0.75}$), Precision (P) and Normalized Precision ($P_{norm}$). Scores are interpreted within the stated dataset protocol, metric and comparator list.

For runtime comparison, executable methods are remeasured on the same hardware/software environment; FPS excludes data loading. SpecTrack-L384 is the accuracy-oriented variant, while B224 and T224 show practical accuracy--speed operating points.

\paragraph{Spectral curve interpretation}
Qualitative curves use discrete recorded bands and benchmark-provided digital-number or normalized-intensity scales. They are reported as template--target--background separability evidence for tracking, with ROI selection and uncertainty handled by the dataset protocol when available.

\subsection{Comparison with Published Trackers}
We evaluate SpecTrack within each benchmark protocol in Tab.~\ref{tab:protocol}; reported numbers remain benchmark-specific.

\begin{figure}[!t]
    \centering
    \includegraphics[width=1\linewidth]{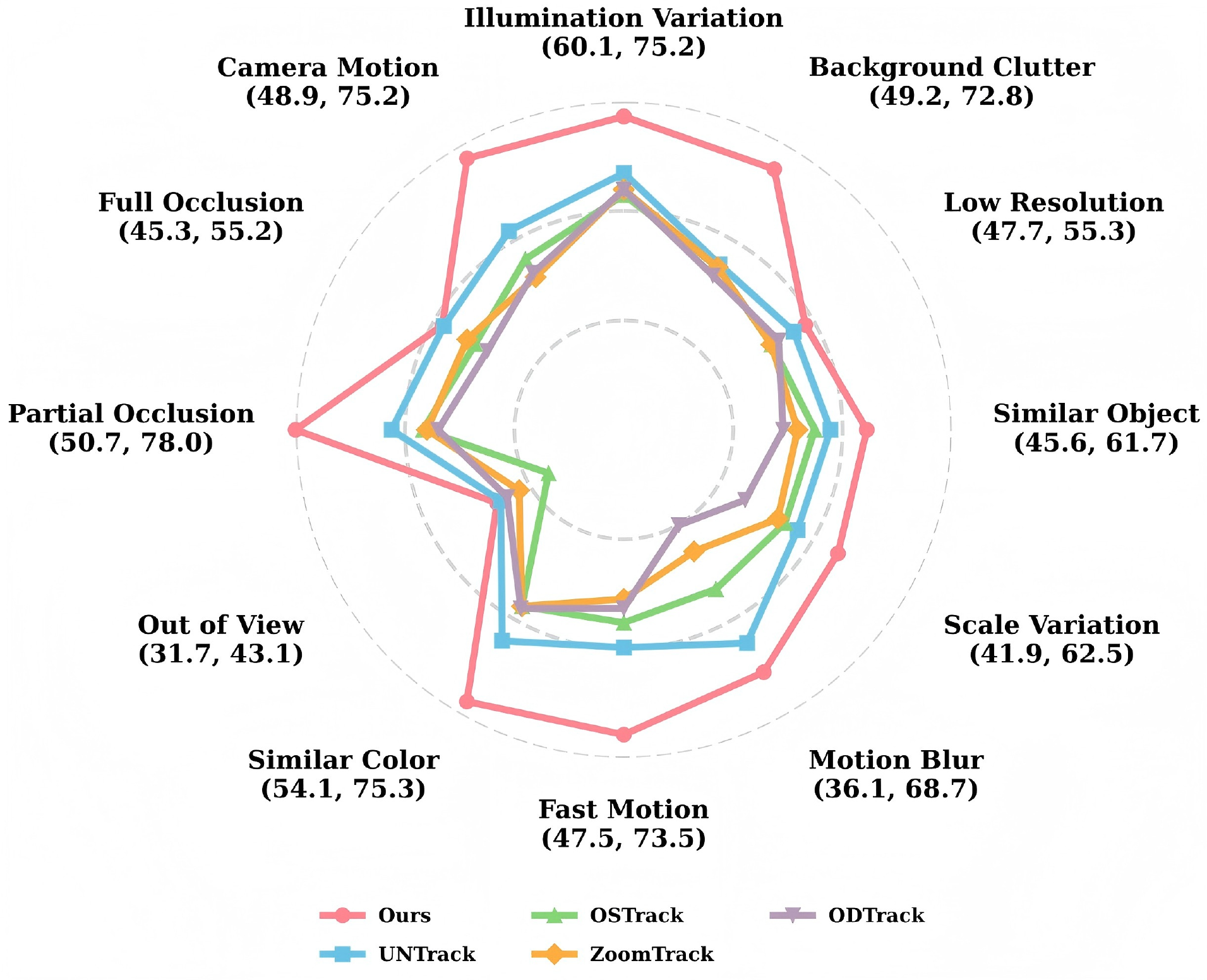}
    \caption{Attribute-level AUC comparison between SpecTrack and published trackers on the MUST dataset. The values in parentheses denote the strongest listed baseline AUC and SpecTrack AUC, respectively.}
    \label{fig:spider}
\end{figure}

\begin{figure}[!t]
    \centering
    \includegraphics[width=1\linewidth]{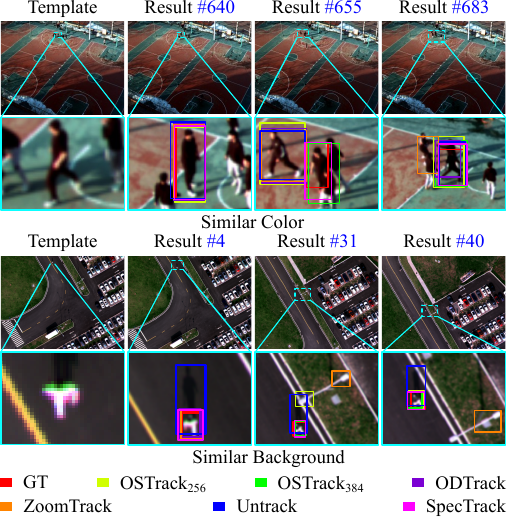}
    \caption{Qualitative tracking examples on two representative MUST sequences with multiple challenge attributes. The predicted boxes illustrate localization behavior under similar color appearance, cluttered background and challenging target background separation.}
    \label{fig:example}
\end{figure}

\paragraph{Evaluation on MUST dataset}
On MUST (Tab.~\ref{tab:MUST_dataset}), SpecTrack-L384 is reported as the accuracy-upper-bound variant: it achieves 65.2\% AUC, 79.8\% $\mathrm{SR}_{0.5}$ and 61.0\% $\mathrm{SR}_{0.75}$, improving over UNTrack by 5.5, 4.0 and 14.8 points. The balanced SpecTrack-B224 variant is the practical default, reaching 62.4\% AUC at 43.7 FPS, which is 2.7 AUC points above UNTrack while running faster under the unified runtime. Thus, the MUST improvement is not only represented by the expensive L384 setting. Fig.~\ref{fig:spider} and Fig.~\ref{fig:example} show gains across challenge categories and similar-background scenes.

\begin{table}[h]
    \centering
    \caption{Comparison with published trackers on the MUST dataset. The best and second best results are marked in bold and underline, respectively. All rerun FPS values are measured under the same runtime environment; ``--'' denotes that no comparable speed is available.}
    \label{tab:MUST_dataset}
    \fontsize{8pt}{9pt}\selectfont
    \begin{tabular*}{\columnwidth}{@{\extracolsep{\fill}}l c c c c c c@{}}
        \toprule
        Method & AUC & $\text{SR}_{0.5}$ & $\text{SR}_{0.75}$ & $\text{Pre}$ & $\text{Pre}_N$ & FPS \\
        \midrule
        OSTrack$_{256}$ \cite{Ye2022OSTrack} & 52.3 & 64.8 & 36.3 & 67.5 & 65.2 &  40.4 \\
        OSTrack$_{384}$ \cite{Ye2022OSTrack} & 55.1 & 69.6 & 44.1 & 73.3 & 68.8 & 21.1 \\
        ODTrack \cite{zheng2024odtrack} & 52.7 & 67.3 & 38.8 & 71.3 & 66.2 & 32.0 \\
        ZoomTrack \cite{kou2023zoomtrack}  & 53.3 & 67.7 & 40.4 & 72.3 & 66.7 & 37.5 \\
        UNTrack \cite{qin_must_2025} & {59.7} & {75.8} & {46.2} & {79.2} & {74.8} & 38.0 \\
        \midrule
        SpecTrack-T224  & {57.9} & {72.9} & 52.0 & {75.3} & {73.5} & \textbf{80.1} \\
        SpecTrack-B224  & {62.4} & {76.6} & \underline{59.3} & {78.8} & {76.2} & \underline{43.7} \\
        SpecTrack-B384  & \underline{63.7} & \underline{78.4} & {58.8} & \underline{80.8} & \underline{77.8} & 17.2 \\
        SpecTrack-L224  & {63.1} & {77.1} & {58.9} & {79.5} & {76.7} & 19.4 \\
        SpecTrack-L384  & \textbf{65.2} & \textbf{79.8} & \textbf{61.0} & \textbf{82.6} & \textbf{79.5} & 5.9 \\
        \bottomrule
    \end{tabular*}
\end{table}

\paragraph{Evaluation on MSITrack dataset}
We compare SpecTrack with published trackers on the MSITrack benchmark in Tab.~\ref{tab:MSITrack_dataset}. GRM is the strongest listed baseline in AUC and Precision. The accuracy-oriented SpecTrack-L384 increases AUC from 47.8\% to 51.9\% and Precision from 63.0\% to 65.0\%. Below the largest variant, SpecTrack-B224 already reaches 48.4\% AUC and SpecTrack-B384 reaches 49.8\% AUC with a 3.0-point gain in $\text{Pre}_N$ over GRM. This shows that the adaptive design contributes across model scales rather than only through the L384 upper-bound setting.

\begin{table}[!t]
    \centering
    \caption{Comparison with published trackers on the MSITrack dataset. The best and second best results are marked in bold and underline, respectively.}
    \label{tab:MSITrack_dataset}
    \fontsize{8pt}{9pt}\selectfont
    \begin{tabular*}{\columnwidth}{@{\extracolsep{\fill}}l c c c c c@{}}
        \toprule
        Method & AUC & $\text{SR}_{0.5}$ & $\text{SR}_{0.75}$ & $\text{Pre}$ & $\text{Pre}_N$ \\
        \midrule
        OSTrack$_{256}$ \cite{Ye2022OSTrack} & 42.0 & 50.9 & 34.2 & 55.7 & 50.5 \\
        OSTrack$_{384}$ \cite{Ye2022OSTrack} & 43.1 & 52.5 & 36.1 & 56.2 & 52.4 \\
        LMTrack \cite{xu2025less} & 43.2 & 52.4 & 39.7 & 55.2 & 52.5 \\
        ZoomTrack \cite{kou2023zoomtrack}& 44.3 & 54.7 & 35.1 & 59.0 & 54.4 \\
        NeighborTrack \cite{chen2023neighbortrack} & 44.8 & 55.2 & 35.7 & 59.7 & 54.6 \\
        EVPTrack \cite{shi2024explicit} & 46.2 & 55.8 & 43.9 & 58.3 & 56.0 \\
        AQATrack \cite{xie2024autoregressive} & 47.1 & 56.7 & 44.7 & 58.5 & 56.9 \\
        UNTrack \cite{qin_must_2025}& 47.3 & 58.4 & 40.5 & 61.8 & 57.8 \\
        GRM \cite{gao2023generalized} & 47.8 & 58.1 & 40.8 & 63.0 & 57.8 \\
        \midrule
        SpecTrack-T224  & {43.3} & {52.7} & 40.7 & {55.1} & {52.2} \\
        SpecTrack-B224  & {48.4} & {58.6} & {43.3} & {63.2} & {58.2} \\
        SpecTrack-B384  & {49.8} & \underline{61.4} & {45.8} & \underline{63.8} & \underline{60.8} \\
        SpecTrack-L224  & \underline{50.7} & {61.1} & \underline{46.4} & {63.5} & {60.7} \\
        SpecTrack-L384  & \textbf{51.9} & \textbf{62.2} & \textbf{48.9} & \textbf{65.0} & \textbf{62.2} \\
        \bottomrule
    \end{tabular*}
\end{table}

\paragraph{Evaluation on HOTC20 dataset}

We evaluate SpecTrack on the 16 band HOTC20 benchmark in Tab.~\ref{tab:HOT_dataset}. This naming follows the dataset source in Xiong et al.~\cite{xiong_material_MHT_2020} and avoids conflating it with later HOTC releases that use different camera subsets and band counts. SpecTrack-L384 achieves the highest AUC of 72.6\% and is interpreted as the accuracy upper bound of the proposed variants. The balanced SpecTrack-B224 obtains 70.2\% AUC at 43.7 FPS, which is competitive with HotMoE and UNTrack at 70.4\% AUC. We therefore present HOTC20 as a clear accuracy--efficiency trade-off: high-capacity variants improve the accuracy ceiling, while B224 provides the practical operating point.

\begin{table}[h]
    \centering
    \caption{Comparison with published trackers on the HOTC20 dataset. The best and second best results are marked in bold and underline, respectively.}
    \label{tab:HOT_dataset}
    \fontsize{8pt}{9pt}\selectfont
    \begin{tabular*}{\columnwidth}{@{\extracolsep{\fill}} l c c c @{}}
        \toprule
        Method & AUC & Pre & FPS \\
        \midrule
        MHT \cite{xiong_material_MHT_2020} & 58.7 & 88.0 & - \\
        BAENet \cite{li_bae-net_2020} & 60.6 & 87.7 & - \\
        SSTNet \cite{Li_SST-Net_2021} & 62.1 & 90.1 & - \\
        SiamHYPER \cite{Liu_SiamHYPER_2022} & 67.8 & 94.5 & 27.7 \\
        SEENet \cite{Li_SEE-Net} & 65.3 & 93.0 & 13.1 \\
        HANet \cite{yu_hanet_2024} & 68.8 & 94.8 & 21.2 \\
        HotMoE \cite{sun2025hotmoe} & 70.4 & 93.5 & \underline{43.7}\\
        UNTrack \cite{qin_must_2025} & 70.4 & 93.7 & 37.0 \\
        \midrule
        SpecTrack-T224  & {63.8} & {91.2} & \textbf{80.1}  \\
        SpecTrack-B224  & {70.2} & {93.4} & \underline{43.7} \\
        SpecTrack-B384  & {71.2} & \underline{95.0} & {17.2}  \\
        SpecTrack-L224  & \underline{72.4} & \textbf{95.1} & {19.4} \\
        SpecTrack-L384  & \textbf{72.6} & \underline{95.0} & {5.9} \\
        \bottomrule
    \end{tabular*}
\end{table}

\paragraph{RGB generalization on GOT-10k}
We further evaluate SpecTrack on the RGB GOT-10k one-shot protocol to examine architectural generalization beyond MSI/HSI tracking. SpecTrack-L384 achieves 79.3\% AO, 88.7\% $\text{SR}_{0.5}$, and 81.4\% $\text{SR}_{0.75}$, ranking first among the compared published trackers in Tab.~\ref{tab:GOT10k_dataset}. This suggests that the proposed capacity-aware expert design can transfer to generic RGB tracking.

\begin{table}[!t]
    \centering
    \caption{Comparison with published trackers on the GOT-10k dataset under the one shot protocol. The best and second best results are marked in bold and underline, respectively.}
    \label{tab:GOT10k_dataset}
    \fontsize{8pt}{9pt}\selectfont
    \begin{tabular*}{\columnwidth}{@{\extracolsep{\fill}} l c c c @{}}
        \toprule
        Method & AO & $\text{SR}_{0.5}$ & $\text{SR}_{0.75}$ \\
        \midrule
        OSTrack$_{256}$ \cite{Ye2022OSTrack} & 71.0 & 80.4 & 68.2 \\
        OSTrack$_{384}$ \cite{Ye2022OSTrack} & 73.7 & 83.2 & 70.8 \\
        LoRAT \cite{lin2024tracking} & 72.1 & 81.8 & 70.7 \\
        SeqTrack \cite{chen2023seqtrack} & 74.7 & 84.7 & 71.8 \\
        SPMTrack \cite{cai2025spmtrack} & 76.5 & 85.9 & 76.3 \\
        UNTrack \cite{qin_must_2025} & 77.3 & \underline{88.4} & 74.7 \\
        \midrule
        SpecTrack-T224  & {72.9} & {82.2} & {71.0}  \\
        SpecTrack-B224  & {77.5} & {87.1} & 78.2 \\
        SpecTrack-B384  & \underline{78.9} & {87.8} & \underline{79.6}  \\
        SpecTrack-L224  & {78.5} & {87.5} & {79.4} \\
        SpecTrack-L384  & \textbf{79.3} & \textbf{88.7} & \textbf{81.4} \\
        \bottomrule
    \end{tabular*}
\end{table}

\subsection{Ablation Studies}
\label{sec:ablation}

The ablations are organized as diagnostic questions rather than repeated ``full model versus baseline'' comparisons. Each group isolates a different part of the proposed capacity-allocation pipeline: the recorded MSI input, the multispectral input adapter, the routing cues, the latent channel-variation cue, the MoE design, the sparse-routing hyperparameters and the resulting routing behavior. Unless otherwise specified, all ablations use SpecTrack-B224 on MUST with the same input resolution, training split and evaluation protocol as the main comparison.

\begin{table}[!t]
    \centering
    \caption{Logical organization of the ablation study. Each experiment answers a distinct question in the proposed capacity-allocation pipeline.}
    \label{tab:ablation_question_map}
    \scriptsize
    \setlength{\tabcolsep}{2pt}
    \renewcommand{\arraystretch}{1.08}
    \begin{tabularx}{\columnwidth}{@{}L{0.31\columnwidth}L{0.28\columnwidth}Y@{}}
        \toprule
        Question & Evidence & Main role \\
        \midrule
        Do recorded MSI bands help tracking? & Tab.~\ref{tab:ablation_spectral} & Separates input evidence from architecture. \\
        How should RGB patch weights be expanded to MSI? & Tab.~\ref{tab:channel_expansion} & Chooses a stable multispectral input adapter. \\
        Which prompt cues guide routing? & Tab.~\ref{tab:prompt_ablation} & Tests semantic, spatial and latent channel cues. \\
        Is the latent channel cue meaningful? & Tab.~\ref{tab:latent_order_sensitivity} & Tests order sensitivity and raw-band replacement. \\
        Is the gain more than generic MoE? & Tab.~\ref{tab:moe_prompt_controls} & Separates capacity ordering, prompt routing and shared context. \\
        What sparse-routing setting is practical? & Tabs.~\ref{tab:ablation_hyperparam}--\ref{tab:topk_ablation} & Balances accuracy and runtime. \\
        Does the router behave as intended? & Tab.~\ref{tab:routing_statistics} & Diagnoses expert usage, entropy and high-capacity allocation. \\
        \bottomrule
    \end{tabularx}
\end{table}

\subsubsection{Input Evidence: Do Recorded Multispectral Bands Help?}

Tab.~\ref{tab:ablation_spectral} separates the contribution of the input modality from the contribution of the proposed architecture. Both the uniform baseline and SpecTrack are evaluated with pseudo-color RGB projections and real MSI inputs.

\begin{table}[!t]
    \centering
    \caption{Ablation study on input modalities. RGB denotes pseudo-color images synthesized from MSI frames, while MSI uses the recorded multispectral input.}
    \label{tab:ablation_spectral}
    \fontsize{8pt}{9pt}\selectfont
    \begin{tabular*}{\columnwidth}{@{\extracolsep{\fill}} l c c c @{}}
        \toprule
        Model & Input Data & AUC & Pre \\
        \midrule
        Baseline & RGB & 55.2 & 71.4 \\
        Baseline & MSI & 59.1 & 76.5 \\
        \midrule
        SpecTrack & RGB & 56.3 & 73.5 \\
        SpecTrack & MSI & \textbf{62.4} & \textbf{78.8} \\
        \bottomrule
    \end{tabular*}
\end{table}

Changing the baseline input from pseudo-RGB to MSI raises AUC from 55.2 to 59.1, showing that recorded bands provide useful target--background evidence even without the proposed adaptive modules. SpecTrack further benefits from MSI input, improving from 56.3 to 62.4 AUC. Thus, the following ablations are conducted in the MSI setting, where the proposed routing cues have meaningful recorded-band-induced variation to exploit.

\subsubsection{Input Adapter: How Should RGB Weights Be Expanded to MSI?}

Tab.~\ref{tab:channel_expansion} examines how the RGB patch embedding is adapted to eight-band MSI. This experiment does not test the MoE mechanism; it controls the multispectral initialization before routing is applied. \textit{Insert halfcopy} gives the best AUC and is used as the default for MUST and MSITrack. HOTC20 uses the interpolation fallback because its 16 bands cover a narrower 470--620\,nm range.

\begin{table}[!t]
    \centering
    \caption{Ablation study on channel expansion strategies in the eight-band MSI setting. $\Delta$ denotes the AUC difference relative to the \textit{copy} baseline.}
    \label{tab:channel_expansion}
    \begingroup
    \fontsize{7.3pt}{8.3pt}\selectfont
    \setlength{\tabcolsep}{2pt}
    \renewcommand{\arraystretch}{1.08}
    \begin{tabularx}{\columnwidth}{@{}l>{\raggedright\arraybackslash}X>{\raggedright\arraybackslash}Xcc@{}}
        \toprule
        Strategy & Weight construction & Target scenario & AUC & $\Delta$ \\
        \midrule
        copy & Duplicate RGB to 8 channels & Naive fill & 61.2 & -- \\
        halfcopy & Duplicate RGB, scaled by $1/2$ & Activation control & 61.8 & +0.6 \\
        random & 3 RGB channels $+$ random init & No band metadata & 60.3 & -0.9 \\
        insert & Proximity to RGB assignment & Known band centers & 62.1 & +0.9 \\
        insert halfcopy & Proximity assignment, scaled by $1/2$ & Default & \textbf{62.4} & \textbf{+1.2} \\
        \bottomrule
    \end{tabularx}
    \endgroup
\end{table}

\subsubsection{Routing Cues: Can Spectral--Spatial Prompts Guide Expert Selection?}

Tab.~\ref{tab:prompt_ablation} focuses on the router input. Semantic-only routing reaches 60.9 AUC. Adding the spatial boundary prompt improves AUC to 61.5, and adding the latent channel-variation prompt improves AUC to 61.7. Their fusion reaches 62.4 AUC, indicating that the two prompts are complementary: the spatial cue emphasizes boundary and texture changes, while the latent channel cue summarizes recorded-band-induced feature variation.

\begin{table}[!t]
    \centering
    \caption{Effect of routing prompts on MUST with SpecTrack-B224. The table isolates the information used by the router while keeping the backbone and expert setting fixed.}
    \label{tab:prompt_ablation}
    \fontsize{8pt}{9pt}\selectfont
    \begin{tabular*}{\columnwidth}{@{\extracolsep{\fill}}lccc@{}}
        \toprule
        Prompt type & AUC & Pre & FPS \\
        \midrule
        Semantic only & 60.9 & 76.7 & 45.8 \\
        Semantic + spatial prompt & 61.5 & 77.9 & 41.9 \\
        Semantic + latent channel-variation prompt & 61.7 & 77.8 & 43.5 \\
        Semantic + spatial + latent channel-variation & \textbf{62.4} & \textbf{78.8} & 43.7 \\
        \bottomrule
    \end{tabular*}
\end{table}

\subsubsection{Latent Cue: Is Channel Order Merely an Unordered Statistic?}

Tab.~\ref{tab:latent_order_sensitivity} tests whether the latent channel-variation cue depends on the learned latent-channel organization. During inference, the trained backbone, expert parameters, semantic descriptor, spatial prompt and expert inputs are kept fixed; only the channel order used by the finite-difference operation in Eq.~\ref{eq:channel_diff} is perturbed. The reverse order preserves a structured but inverted neighborhood, random permutations destroy most latent-channel adjacency relations and the raw-band baseline replaces the latent cue with finite differences along the recorded band order.

\begin{table*}[!t]
    \centering
    \caption{Latent-channel order sensitivity of the channel-variation prompt on MUST with SpecTrack-B224. Accuracy values follow the same MUST reporting protocol as the main ablation setting; deltas are computed relative to the unperturbed identity model. High-capacity statistics refer to the logged usage and gate mass associated with the stronger experts E3 and E4. Usage L1 and Gate L1 measure distribution shifts from the identity routing behavior.}
    \label{tab:latent_order_sensitivity}
    \begin{threeparttable}
    \fontsize{8pt}{9pt}\selectfont
    \renewcommand{\arraystretch}{1.06}
    \begin{adjustbox}{max width=\textwidth}
    \begin{tabular}{@{}llrrrrrrrrrr@{}}
        \toprule
        Seed & Setting & AUC & $\Delta$AUC & Pre & $\Delta$Pre & High-cap. usage & $\Delta$usage & High-cap. gate & $\Delta$gate & Usage L1 & Gate L1 \\
        \midrule
        identity & Full SpecTrack-B224 & 62.4 & 0.0 & 78.8 & 0.0 & 0.550 & 0.000 & 0.196 & 0.000 & 0.000 & 0.000 \\
        reverse & Reverse latent order & 61.5 & -0.9 & 77.8 & -1.0 & 0.579 & 0.029 & 0.239 & 0.043 & 0.247 & 0.350 \\
        seed 0 & Random latent permutation & 61.3 & -1.1 & 77.8 & -1.0 & 0.563 & 0.013 & 0.231 & 0.036 & 0.187 & 0.296 \\
        seed 1 & Random latent permutation & 59.9 & -2.5 & 75.8 & -3.0 & 0.558 & 0.009 & 0.203 & 0.007 & 0.296 & 0.357 \\
        seed 2 & Random latent permutation & 61.3 & -1.1 & 77.7 & -1.1 & 0.554 & 0.004 & 0.246 & 0.050 & 0.085 & 0.180 \\
        seed 3 & Random latent permutation & 61.1 & -1.3 & 77.6 & -1.2 & 0.566 & 0.017 & 0.235 & 0.039 & 0.200 & 0.343 \\
        seed 4 & Random latent permutation & 62.3 & -0.1 & 78.7 & -0.1 & 0.588 & 0.038 & 0.256 & 0.060 & 0.178 & 0.266 \\
        raw band & Raw-band prompt baseline & 60.4 & -2.0 & 76.6 & -2.2 & 0.557 & 0.007 & 0.215 & 0.019 & 0.181 & 0.249 \\
        \bottomrule
    \end{tabular}
    \end{adjustbox}
    \begin{tablenotes}
        \footnotesize
        \item Random latent permutation over five seeds: AUC $61.2\pm0.7$, Precision $77.5\pm0.9$, $\Delta$AUC $=-1.2$ and $\Delta$Precision $=-1.3$.
    \end{tablenotes}
    \end{threeparttable}
\end{table*}

Reversing the latent order reduces AUC by 0.9 points, while random permutations reduce AUC by 1.2 points on average. The raw-band prompt baseline decreases AUC by 2.0 points, indicating that direct finite differences along recorded bands do not substitute for the learned latent routing cue. Corrupted orders can increase high-capacity usage or gate mass while lowering tracking accuracy, so the benefit does not come from selecting stronger experts more often; it comes from assigning them according to a better-aligned latent ambiguity cue.

\subsubsection{Expert Design: Is the Gain More Than Generic MoE?}

Tab.~\ref{tab:moe_prompt_controls} compacts the overlapping module, MoE and shared-context controls into a single diagnostic table. The rows answer different questions. Capacity experts alone test whether multiple branches are enough; the semantic router tests whether generic semantic routing is enough; homogeneous MoE-style experts test whether generic expert branching can replace capacity ordering; and the local-only setting tests whether shared global context is needed under sparse routing.

\begin{table}[!t]
    \centering
    \caption{Compact MoE/prompt controls on MUST under the same SpecTrack-B224 protocol. $\Delta$ denotes the AUC change relative to the uniform backbone baseline. ``Fused prompt'' denotes semantic, spatial and latent channel-variation cues.}
    \label{tab:moe_prompt_controls}
    \fontsize{8pt}{9pt}\selectfont
    \setlength{\tabcolsep}{3pt}
    \renewcommand{\arraystretch}{1.08}
    \begin{tabularx}{\columnwidth}{@{}cXcc@{}}
        \toprule
        \# & Variant / controlled question & AUC & $\Delta$ \\
        \midrule
        1 & Uniform backbone baseline & 59.1 & 0.0 \\
        2 & Capacity experts only: are multiple capacity branches sufficient? & 60.9 & +1.8 \\
        3 & Capacity experts + semantic router: is generic semantic routing sufficient? & 60.9 & +1.8 \\
        4 & Homogeneous MoE + fused prompt: is generic MoE sufficient? & 61.4 & +2.3 \\
        5 & Capacity experts + fused prompt, w/o shared global context & 61.5 & +2.4 \\
        6 & \textbf{SpecTrack full} & \textbf{62.4} & \textbf{+3.3} \\
        \bottomrule
    \end{tabularx}
\end{table}

The compact controls show that the final improvement is not explained by MoE-like branches alone. Capacity experts without prompt routing improve the uniform baseline but remain below the full model. Semantic-only routing gives no additional gain over capacity experts alone, suggesting that MSI tracking requires more than generic semantic descriptors. Homogeneous experts under the fused prompt are also weaker than capacity-ordered experts, and removing the Shared Global Expert reduces the AUC from 62.4 to 61.5. Therefore, the gain comes from the combination of capacity ordering, multispectral prompt-guided routing and shared global context.

\begin{figure}[!t]
    \centering
    \includegraphics[width=1\linewidth]{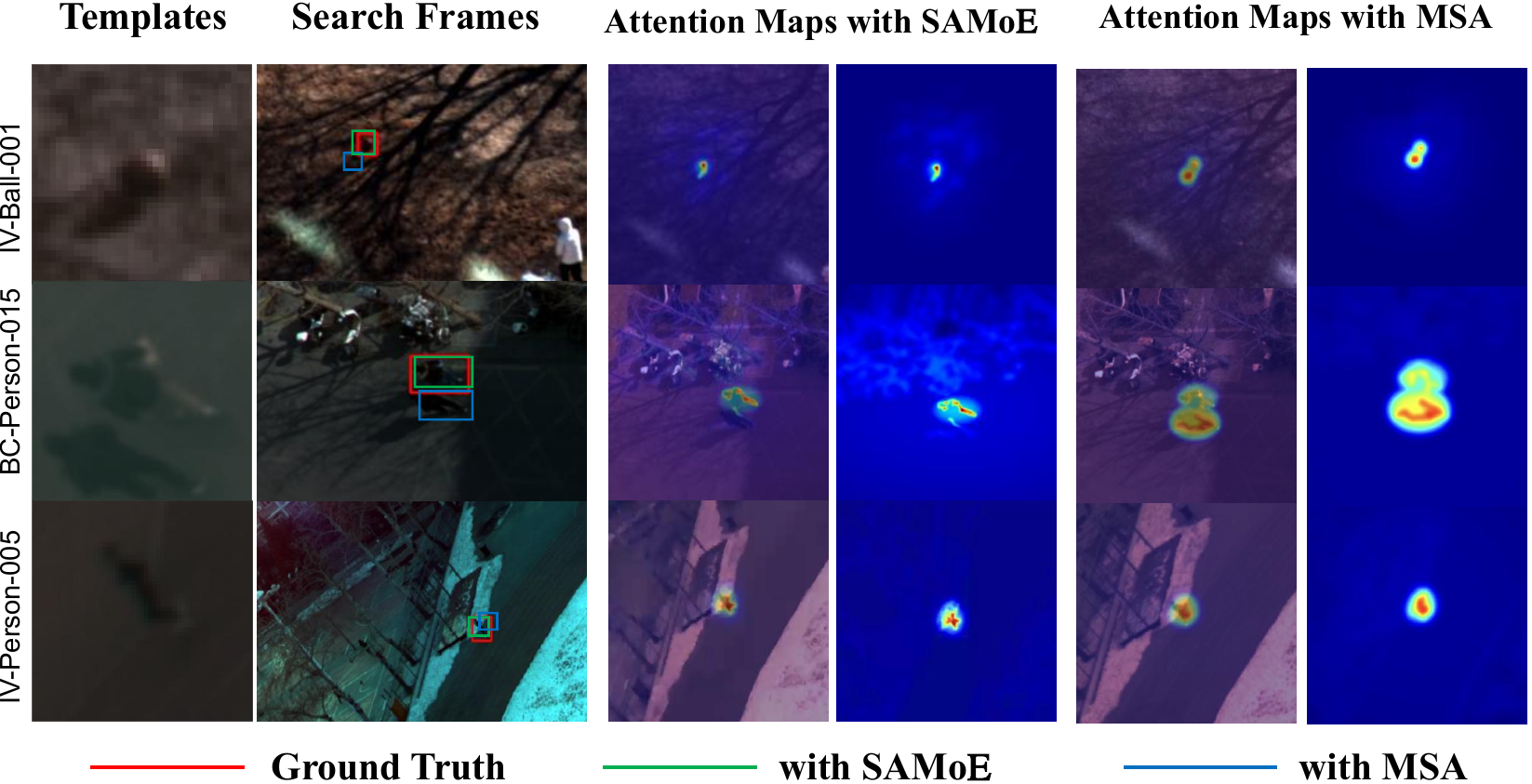}
    \caption{Attention map comparison between the uniform processing MSA baseline and the adaptive SAMoE block. The SAMoE-enhanced block denotes the adaptive expert block used by SpecTrack, whereas MSA denotes the uniform baseline. The adaptive block produces more target-concentrated responses in cluttered regions, supporting the capacity allocation interpretation.}
    \label{fig:samoe_attention}
\end{figure}

\subsubsection{Sparse Routing: What Accuracy--Efficiency Setting Is Practical?}

After the mechanism is fixed, Tabs.~\ref{tab:ablation_hyperparam} and~\ref{tab:topk_ablation} tune the number of experts and the routing sparsity. These experiments answer a deployment question rather than another component-removal question.

\begin{table}[!t]
    \centering
    \caption{Ablation study on the number of experts ($N$) with the activated experts fixed to $k=2$. The default setting is $N=4, k=2$.}
    \label{tab:ablation_hyperparam}
    \fontsize{8pt}{9pt}\selectfont
    \begin{tabular*}{\columnwidth}{@{\extracolsep{\fill}} c c c c c @{}}
        \toprule
        $N$ (Total) & $k$ (Top $k$) & AUC & Pre & FPS \\
        \midrule
        3 & 2 & 62.2 & 78.6 & \textbf{45.4} \\
        4 & 2 & 62.4 & \textbf{78.8} & 43.7 \\
        5 & 2 & \textbf{62.5} & \textbf{78.8} & 39.7 \\
        6 & 2 & 62.1 & 78.4 & 35.2 \\
        \bottomrule
    \end{tabular*}
\end{table}

\begin{table}[!t]
    \centering
    \caption{Effect of Top-$k$ routing with $N=4$ experts on MUST.}
    \label{tab:topk_ablation}
    \fontsize{8pt}{9pt}\selectfont
    \begin{tabular*}{\columnwidth}{@{\extracolsep{\fill}}cccc@{}}
        \toprule
        $N$ & $k$ & AUC & FPS \\
        \midrule
        4 & 1 & 61.1 & \textbf{52.4} \\
        4 & 2 & \textbf{62.4} & 43.7 \\
        4 & 3 & 62.3 & 33.6 \\
        4 & 4 & 62.1 & 27.3 \\
        \bottomrule
    \end{tabular*}
\end{table}

With $k=2$, increasing $N$ from 3 to 5 gives only a marginal accuracy gain, while $N=6$ reduces both accuracy and speed. With $N=4$, Top-$k=1$ is fastest but loses accuracy, whereas $k=3$ and $k=4$ increase computation without improving AUC. We therefore use $N=4,k=2$ as the default accuracy--efficiency point.

\subsubsection{Routing Diagnostics: Does the Router Behave as Intended?}

The previous tables measure accuracy. Tab.~\ref{tab:routing_statistics} instead examines routing behavior at the search-sample level. Usage denotes the percentage of selected expert slots and $H_p$ is the dense-router entropy before Top-$k$ selection.

\begin{table}[!t]
    \centering
    \caption{Routing statistics of SAMoE on the MUST test split. Usage denotes Top-$k=2$ selected-slot frequency. $H_p$ is computed from dense-router probabilities before Top-$k$.}
    \label{tab:routing_statistics}
    \fontsize{8pt}{9pt}\selectfont
    \begin{tabular*}{\columnwidth}{@{\extracolsep{\fill}}lccccc@{}}
        \toprule
        Setting & E1 & E2 & E3 & E4 & $H_p$ \\
        \midrule
        Default SAMoE & 23.9 & 21.2 & 24.3 & 30.6 & 0.353 \\
        w/o latent channel prompt & 23.5 & 26.0 & 24.7 & 25.8 & 0.431 \\
        w/o shared expert & 23.3 & 23.2 & 26.0 & 27.5 & 0.423 \\
        \bottomrule
    \end{tabular*}
\end{table}

The full router does not collapse to a single expert: all four experts are used, and the stronger experts E3 and E4 receive 54.9\% of the selected slots. Removing the latent channel prompt reduces E4 usage from 30.6\% to 25.8\% and increases entropy, indicating less confident capacity assignment. Removing the shared expert also weakens the high-capacity preference. Logged samples show a Spearman correlation of 0.60 between latent channel-variation strength and the pre-renormalization retained mass of E3+E4, supporting the link between embedded spectral--spatial ambiguity and high-capacity expert allocation.

\section{Discussion and Limitations}
\label{sec:discussion}

\paragraph{Remote-sensing interpretation.}
SpecTrack uses recorded multispectral observations as routing evidence for tracking, rather than as calibrated material spectra. The Spectral Prompt Router learns how latent channel variation, boundary response, and semantic context jointly indicate the difficulty of a search region. This design is practical for benchmark tracking data where complete radiometric calibration metadata are often unavailable. When calibrated reflectance, bandpass information, and sensor response functions are available, the routing cues could be extended toward more sensor-aware or material-aware tracking designs.

\begin{figure}[!t]
    \centering
    \includegraphics[width= \linewidth]{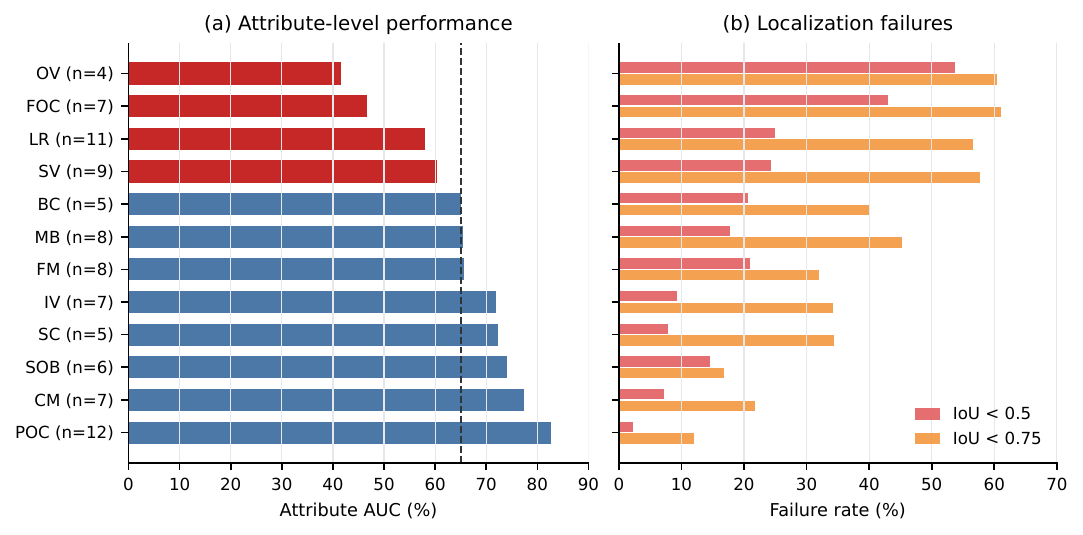}
    \caption{Failure-oriented attribute analysis of SpecTrack-B224 on MUST. Attributes are sorted by AUC in ascending order. Red bars mark the four weakest attributes, and the dashed vertical line denotes the unweighted attribute mean. The right panel reports the percentage of frames whose predicted box falls below IoU thresholds 0.5 and 0.75. Out-of-view, full-occlusion, low-resolution, and scale-variation cases remain the dominant failure modes.}
    \label{fig:spectrack_failure_attribute_analysis}
\end{figure}

\paragraph{Failure-oriented analysis.}
Fig.~\ref{fig:spectrack_failure_attribute_analysis} summarizes the attribute-level limitations of SpecTrack-B224 on MUST. The weakest cases are out-of-view, full occlusion, low resolution, and scale variation. These failures are consistent with the design scope of SpecTrack: adaptive expert routing can allocate stronger local spectral--spatial capacity when useful target evidence remains in the search region, but it cannot reliably recover a target that is absent from the crop or represented by only a few mixed pixels.

These observations suggest that future work should combine adaptive capacity allocation with global re-detection, longer temporal memory, uncertainty-aware template update, and sensor-calibrated prompts when richer radiometric metadata are available.

\section{Conclusion}
\label{sec:conclusion}

This paper presented SpecTrack, a spectral--spatial complexity-aware tracker that formulates MSI/HSI tracking as search-region-level adaptive capacity allocation. By combining capacity-ordered experts, prompt-guided sparse routing, and shared global context, SpecTrack adapts the modeling capacity to the ambiguity of each search region instead of processing all regions through a fixed-capacity path.

Experiments on MUST, MSITrack, and HOTC20 demonstrate that SpecTrack achieves a favorable accuracy--efficiency trade-off, with SpecTrack-B224 serving as the balanced default and SpecTrack-L384 reporting the accuracy-oriented setting. Mechanism-aligned ablations and routing diagnostics further support the roles of multispectral input, spectral--spatial prompts, capacity-ordered experts, and shared global context. The additional GOT-10k evaluation indicates RGB-domain architectural generalization. Remaining limitations mainly arise from out-of-view, full-occlusion, low-resolution, and scale-variation cases, suggesting future directions such as global re-detection, longer temporal memory, and sensor-aware calibration.

\bibliographystyle{elsarticle-num-nourl}
\bibliography{references2}
\end{document}